\documentclass[conference]{IEEEtran}
\IEEEoverridecommandlockouts
\usepackage{cite}
\usepackage{amsmath,amssymb,amsfonts}
\usepackage{algorithmic}
\usepackage{graphicx}
\usepackage{textcomp}
\usepackage{xcolor}
\usepackage{glossaries}
\usepackage{float}
\usepackage{tikz}
\usepackage{booktabs}
\usetikzlibrary{arrows.meta, positioning, shapes.geometric, backgrounds, fit}

\makeglossaries

\def\BibTeX{{\rm B\kern-.05em{\sc i\kern-.025em b}\kern-.08em
    T\kern-.1667em\lower.7ex\hbox{E}\kern-.125emX}}

\newacronym{VLA}{VLA}{Vision-Language-Action}
\newacronym{LSTM}{LSTM}{Long Short-Term Memory}
\newacronym{SLR}{SLR}{Sign Language Recognition}
\newacronym{SLT}{SLT}{Sign Language Translation}
\newacronym{ASL}{ASL}{American Sign Language}
\newacronym{HRI}{HRI}{Human-Robot Interaction}
\newacronym{LLM}{LLM}{Large Language Model}
\newacronym{LIBERO}{LIBERO}{LIfelong learning BEchmark on RObot manipulation tasks}

\begin{document}

\title{SignVLA: Real-Time Sign Language-Guided Robotic Manipulation via Attention LSTM and Vision-Language-Action Models}

\author{

    \IEEEauthorblockN{
    Ningwei Bai$^{1}$,
    Xinyu Tan$^{1}$,
    Harry Gardner$^{1}$,
    Zhengyang Zhong$^{1}$,
    Liuhaichen Yang$^{1}$ \\
    Luoyu Zhang$^{1}$,
    Zhekai Duan$^{1}$,
    Monkgogi Galeitsiwe$^{1}$,
    Zezhi Tang$^{1*}$}

\thanks{$^1$ Department of Computer Science, University College London.}
\thanks{Correspondence: zezhi.tang@ucl.ac.uk}

}

\maketitle

\begin{abstract}
Vision-Language-Action (VLA) models enable robots to execute manipulation tasks from natural-language instructions grounded in visual observations. However, existing VLA interfaces primarily rely on speech or text input, limiting accessibility for deaf, hard-of-hearing, and speech-impaired users. We present SignVLA, a real-time sign-language-guided VLA framework for accessible human--robot interaction. The system introduces a modular sign-to-text interface that converts visual sign gestures into semantic instructions compatible with downstream VLA policies. Given video streams, SignVLA extracts hand landmark features and employs an attention-enhanced Long Short-Term Memory (LSTM) network to capture temporal gesture dynamics for alphabet- and command-level sign recognition. A temporal stabilization module further improves prediction consistency in real-time interaction settings.The generated instruction sequence is then passed to a downstream VLA policy for sign-conditioned robotic manipulation. Experimental results demonstrate stable real-time sign recognition and successful execution of manipulation tasks driven by sign-language inputs. Our findings suggest that lightweight temporal sign recognition can serve as an effective and practical accessibility layer for multimodal embodied intelligence.
\end{abstract}


\section{Introduction}

Vision-Language-Action (VLA) models have recently emerged as a promising paradigm for generalist robotic manipulation. By jointly conditioning on visual observations and language instructions, such models can ground semantic goals in physical scenes and generate executable robot actions. Representative systems, including PaLM-E, RT-2, OpenVLA, and GR00T N1, demonstrate that large-scale vision-language pretraining and robot demonstration data can substantially improve semantic generalization, cross-task transfer, and embodied decision-making \cite{driess2023palme,zitkovich2023rt2,kim2025openvla,bjorck2025groot,10532060,martinchan}. Despite this progress, the dominant interaction assumption remains narrow: users are typically expected to provide instructions through spoken or written language. This assumption restricts the accessibility of VLA-based robots for deaf, hard-of-hearing, and speech-impaired users, and it also limits deployment in environments where speech input is unreliable, noisy, or socially inappropriate.

Traditional robotic control provides a complementary perspective. Disturbance-observer and iterative-learning methods emphasize tracking accuracy and uncertainty rejection, while reinforcement-learning-enhanced robust control and formation-control schemes improve adaptation or coordination under structured dynamics \cite{tang2019disturbance,tang2026disturbance,tang2024reinforcement,bai2025deep,onuoha2024discrete}. Vision-driven manipulation pipelines further show that deep perception can support task-specific robot operation \cite{zhao2024realtime}. Compared with these model- or domain-structured systems, VLA models move the interface toward semantic language grounding and cross-task transfer, but they still depend on reliable front-end interfaces that convert human intent into stable executable commands.

Beyond control and policy learning, effective human-robot interaction also depends on accessible and reliable input modalities. Existing VLA systems primarily rely on speech or text instructions, motivating exploration of alternative communication interfaces for users in speech-limited or noisy environments. Among these modalities, sign language provides a natural visual communication channel for deaf and hard-of-hearing users and has attracted increasing attention in computer vision and sequence modeling research.

Recent sign language recognition and translation systems based on CNNs, Transformers, and multimodal sequence modeling have demonstrated strong performance on large-vocabulary continuous sign-language benchmarks\cite{yang2024signformerneededgeai,camgoz2020sign,hwang2025efficientsignlanguagetranslation,fang2025signllmsignlanguageproduction,zhou2023glossfreesignlanguagetranslation}. These approaches can translate complex gesture sequences into natural-language sentences with high semantic richness. However, such models typically rely on large-scale annotated datasets, computationally intensive training procedures, and high-capacity inference backbones. Their relatively high latency and hardware requirements make real-time deployment on resource-constrained robotic platforms challenging. 

However, integrating sign input into VLA-based robotic control is non-trivial. Unlike written instructions, signs are continuous spatio-temporal visual signals involving hand shape, finger articulation, hand position, motion trajectory, and temporal context. These properties create a modality mismatch between visual sign streams and the discrete token-based instruction interface typically expected by VLA policies. Moreover, robotic execution is sensitive to recognition instability, transient frame-level errors can produce command flickering, incorrect task selection, or unsafe downstream actions. Therefore, a practical sign-guided robotic system must not only recognize gestures accurately, but also convert noisy visual observations into stable semantic commands with low latency.

To address this gap, we propose SignVLA, a real-time sign-language-guided \gls{VLA} framework. A lightweight sign-to-text interface sits between visual gesture perception and robot manipulation: hand landmark features are extracted from live video using MediaPipe~\cite{zhang2020mediapipe}, processed by an attention-enhanced \gls{LSTM}~\cite{hochreiter1997lstm,bahdanau2015neural} that captures temporal gesture dynamics, and stabilized through temporal filtering before being converted into natural-language task instructions for the downstream \gls{VLA} policy. This design avoids modifying the \gls{VLA} backbone while introducing sign language as a low-latency accessible input modality. The paper contributes a lightweight real-time sign-to-text recognition module, an integrated sign-language-guided \gls{VLA} interaction system that bridges continuous visual gestures and discrete robotic instructions, and experimental validation showing that temporal smoothing and semantic command synthesis improve instruction stability for sign-conditioned robotic manipulation.

\section{Related Work}

Early \gls{SLR} relied on hand-crafted features, but neural approaches now dominate. Camgoz~\emph{et al.}~\cite{camgoz2020sign} introduced a Transformer jointly learning recognition and translation under gloss supervision, while gloss-free approaches~\cite{zhou2023gloss} remove the need for costly gloss annotations. The \gls{ASL} Citizen dataset~\cite{desai2023asl} provides a large community-sourced benchmark suitable for training a domain-specific manipulation classifier. In the \gls{VLA} space, RT-2~\cite{zitkovich2023rt2} demonstrated that internet-scale pretraining enables generalisation to novel instructions; OpenVLA~\cite{kim2025openvla} open-sourced this paradigm; and GR00T N1~\cite{bjorck2025groot} further separates semantic grounding from motor control via a Diffusion Transformer. To our knowledge, no prior work has directly connected sign language recognition to a \gls{VLA} backend for physical robot manipulation.

\section{Method}

\begin{figure}[t]
\centering
\resizebox{0.7\columnwidth}{!}{%
\begin{tikzpicture}[
  every node/.style={font=\small},
  box/.style={draw, rounded corners=2pt, minimum width=4.2cm, minimum height=0.72cm,
              align=center, inner sep=4pt},
  a1/.style={box, fill=blue!10, draw=blue!50},
  lang/.style={box, fill=yellow!18, draw=yellow!65!black},
  rb/.style={box, fill=gray!15, draw=gray!60},
  arr/.style={->, >=Stealth, semithick},
]

\node[a1]   (webcam) at (0, 7.0) {Webcam\\{\tiny live video, RGB}};
\node[a1]   (mp)     at (0, 5.6) {MediaPipe Hands\\{\tiny 544-dim landmark features / frame}};
\node[a1]   (lstm)   at (0, 4.2) {Attention \gls{LSTM}\\{\tiny 33-sign \gls{ASL} vocabulary, $T{=}32$ frames}};
\node[a1]   (buf)    at (0, 2.8) {Temporal Stability Buffer\\{\tiny 6-frame window, cooldown\,=\,8 frames}};
\node[lang] (ds4)    at (0, 1.4) {LLMs\\{\tiny gloss sequence $\to$ natural language instruction}};
\node[rb]   (groot)  at (0, 0.0) {GR00T N1 \gls{VLA}\\{\tiny VLM + Diffusion Transformer}};
\node[rb]   (franka) at (0,-1.4) {Franka Emika Panda\\{\tiny 7-DOF, 100\,Hz planner / 1\,kHz servo}};

\draw[arr] (webcam) -- (mp);
\draw[arr] (mp)     -- (lstm);
\draw[arr] (lstm)   -- (buf);
\draw[arr] (buf)    -- (ds4);
\draw[arr] (ds4)    -- (groot);
\draw[arr] (groot)  -- (franka);

\begin{scope}[on background layer]
  \node[draw=blue!25, fill=blue!5, rounded corners=6pt,
        fit=(webcam)(mp)(lstm)(buf), inner sep=8pt] {};
  \node[draw=yellow!45!black, fill=yellow!5, rounded corners=6pt,
        fit=(ds4), inner sep=8pt] {};
  \node[draw=gray!35, fill=gray!8, rounded corners=6pt,
        fit=(groot)(franka), inner sep=8pt] {};
\end{scope}

\end{tikzpicture}%
}
\caption{SignVLA system architecture. \textcolor{blue!60!black}{Blue}: sign perception (MediaPipe + Attention \gls{LSTM} + temporal buffer). \textcolor{yellow!65!black}{Amber}: language conversion (LLMs). \textcolor{gray!60!black}{Gray}: \gls{VLA} policy and robot execution.}
\label{fig:architecture}
\end{figure}
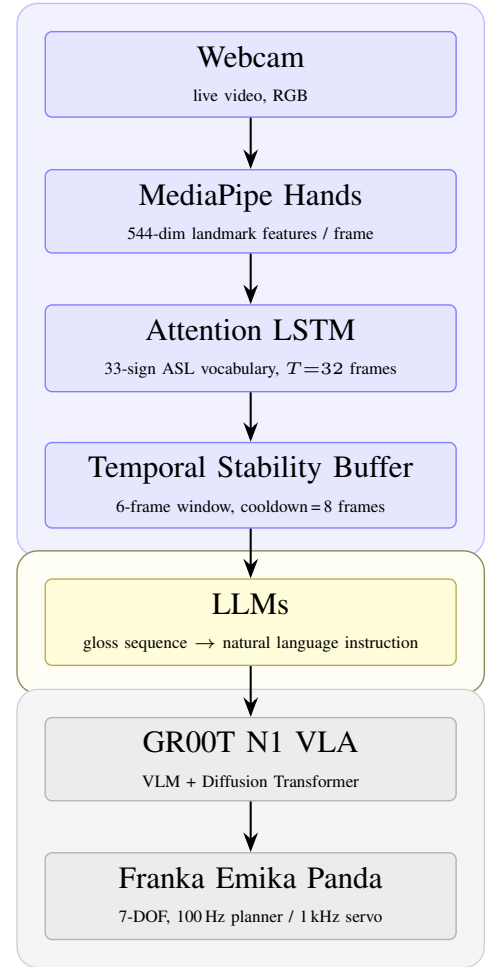

The SignVLA pipeline is shown in Fig.~\ref{fig:architecture}. Sign perception extracts hand landmarks and classifies signs in real time; a temporal stability buffer filters the output before a large language model converts the gloss sequence to a natural-language instruction; the instruction is then consumed by GR00T N1 to drive the Franka arm. 

\subsection{Sign Language Recognition and Translation}

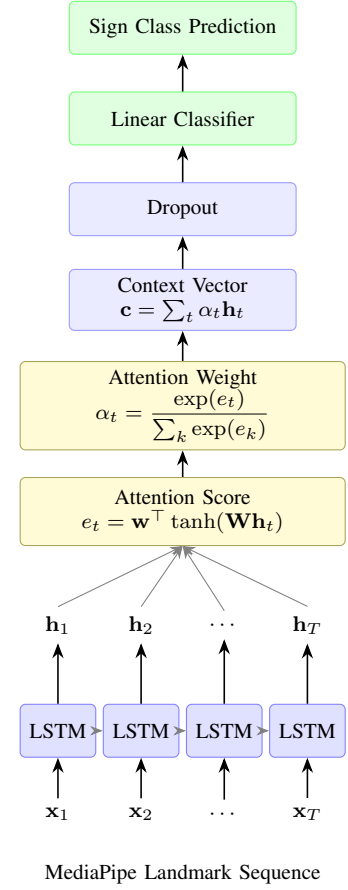
\begin{figure}[t]
\centering
\begin{tikzpicture}[
  every node/.style={font=\small},
  cell/.style={
      draw,
      rounded corners=2pt,
      fill=blue!12,
      draw=blue!50,
      minimum width=1.0cm,
      minimum height=0.7cm,
      align=center,
      inner sep=2pt,
      font=\footnotesize
  },
  box/.style={
      draw,
      rounded corners=2pt,
      fill=blue!8,
      draw=blue!40,
      minimum width=3.0cm,
      minimum height=0.7cm,
      align=center,
      inner sep=3pt,
      font=\footnotesize
  },
  attn/.style={
      draw,
      rounded corners=2pt,
      fill=yellow!20,
      draw=yellow!60!black,
      minimum width=4.3cm,
      minimum height=0.9cm,
      align=center,
      inner sep=3pt,
      font=\footnotesize
  },
  outbox/.style={
      draw,
      rounded corners=2pt,
      fill=green!12,
      draw=green!55,
      minimum width=3.0cm,
      minimum height=0.7cm,
      align=center,
      inner sep=3pt,
      font=\footnotesize
  },
  arr/.style={->, >=Stealth, semithick},
  tarr/.style={->, >=Stealth, thin, gray},
]

\node[font=\footnotesize] at (1.65,-0.8)
{MediaPipe Landmark Sequence};

\foreach \i/\lbl in {
0/$\mathbf{x}_1$,
1/$\mathbf{x}_2$,
2/$\cdots$,
3/$\mathbf{x}_T$
}{
  \node[font=\footnotesize] (in\i) at (\i*1.1,0.0) {\lbl};
}

\foreach \i in {0,1,2,3}{
  \node[cell] (lstm\i) at (\i*1.1,1.1) {LSTM};
}

\foreach \i/\j in {0/1,1/2,2/3}{
  \draw[tarr] (lstm\i.east) -- (lstm\j.west);
}

\foreach \i in {0,1,2,3}{
  \draw[arr] (in\i.north) -- (lstm\i.south);
}

\foreach \i/\lbl in {
0/$\mathbf{h}_1$,
1/$\mathbf{h}_2$,
2/$\cdots$,
3/$\mathbf{h}_T$
}{
  \node[font=\footnotesize] (h\i) at (\i*1.1,2.5) {\lbl};
  \draw[arr] (lstm\i.north) -- (h\i.south);
}

\node[attn] (score) at (1.65,4.0)
{
Attention Score\\
$e_t = \mathbf{w}^\top \tanh(\mathbf{W}\mathbf{h}_t)$
};

\foreach \i in {0,1,2,3}{
  \draw[tarr] (h\i.north) -- (score.south);
}

\node[attn] (softattn) at (1.65,5.4)
{
Attention Weight\\
$\alpha_t = \dfrac{\exp(e_t)}
{\sum_k \exp(e_k)}$
};

\draw[arr] (score.north) -- (softattn.south);

\node[box] (ctx) at (1.65,6.8)
{
Context Vector\\
$\mathbf{c} = \sum_t \alpha_t \mathbf{h}_t$
};

\draw[arr] (softattn.north) -- (ctx.south);

\node[box] (drop) at (1.65,8.0) {Dropout};

\node[outbox] (lin) at (1.65,9.2)
{Linear Classifier};

\node[outbox] (pred) at (1.65,10.4)
{Sign Class Prediction};

\draw[arr] (ctx.north) -- (drop.south);
\draw[arr] (drop.north) -- (lin.south);
\draw[arr] (lin.north) -- (pred.south);

\end{tikzpicture}

\caption{
Architecture of the proposed attention-based \gls{LSTM} framework for isolated sign language classification.
A sequence of MediaPipe landmark features extracted from sign videos is first processed by the temporal \gls{LSTM} encoder to generate hidden representations.
An additive attention mechanism computes temporal attention scores and corresponding attention weights to selectively emphasize discriminative motion segments.
The weighted hidden states are aggregated into a context vector, which is subsequently passed through dropout and a linear classifier for final sign prediction.
}

\label{fig:LSTM_SL}

\end{figure}

To achieve robust real-time sign language detection and translation, an attention-based \gls{LSTM} framework is proposed for isolated sign classification, as shown in Figure~\ref{fig:LSTM_SL}. In order to effectively capture the temporal dynamics and discriminative motion patterns in sign language sequences, an additive attention mechanism is incorporated into the \gls{LSTM} architecture\cite{bahdanau2015neural}.

For dataset construction, 33 commonly used sign language vocabularies were selected from the Microsoft ASL Citizen dataset\cite{desai2023asl}. These selected words mainly consist of frequently used gestures in VLA-based human--robot interaction scenarios to facilitate practical real-time communication and instruction understanding. Each sign video sequence was preprocessed before being used for model training and evaluation.

\begin{figure}[H]
    \centering
    \includegraphics[width=0.6\linewidth]{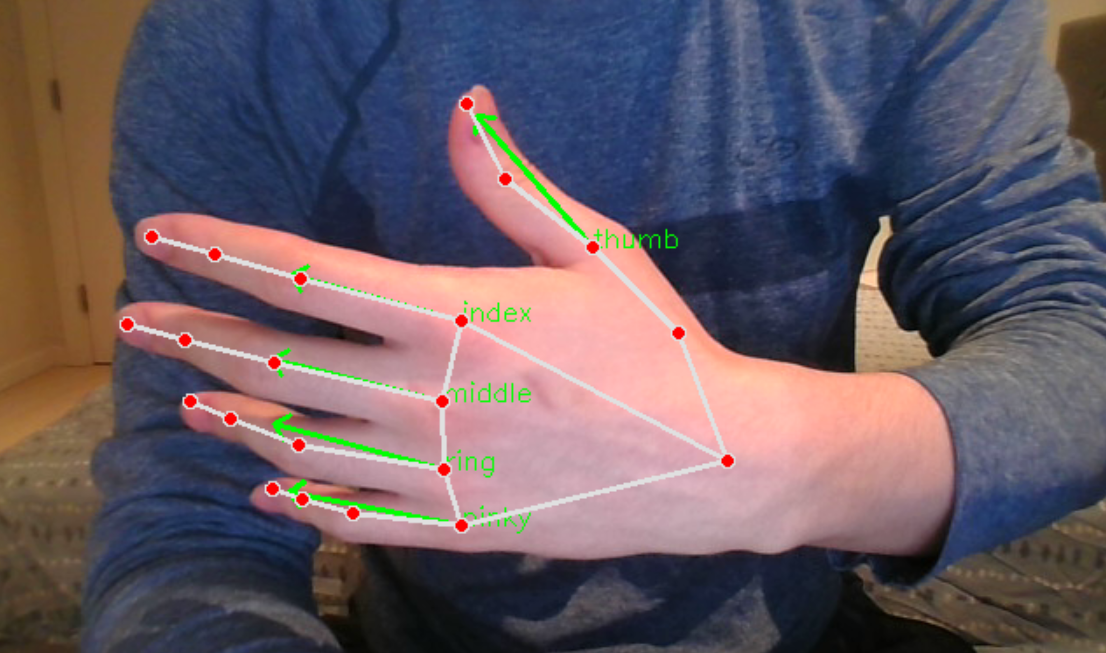}
    \caption{Hand landmark extraction using MediaPipe}
    \label{fig:mediapipe_detect}
\end{figure}

Instead of using raw RGB frames, MediaPipe was employed to extract hands landmark features from sign language videos in real time, as shown in Figure~\ref{fig:mediapipe_detect}.

Given a sequence of MediaPipe landmark features:

\begin{equation}
X = \left\{x_1, x_2, \ldots, x_T \right\},
\quad x_t \in \mathbb{R}^{d}
\end{equation}

where $x_t$ denotes the landmark feature vector extracted at time step $t$, and $T$ represents the temporal sequence length.

The extracted landmark sequences are subsequently fed into the \gls{LSTM} network to model temporal dependencies across the signing sequence. The temporal hidden representation generated by the \gls{LSTM} is defined as:

\begin{equation}
h_t = \mathrm{LSTM}(x_t, h_{t-1})
\end{equation}

where $h_t$ denotes the hidden state at time step $t$.

To further enhance temporal feature aggregation, an additive attention mechanism is applied to the hidden states generated by the \gls{LSTM}. The attention scores are computed using a lightweight multilayer perceptron with nonlinear activation:

\begin{equation}
e_t = \mathbf{W}_2 \tanh(\mathbf{W}_1 h_t + b_1) + b_2
\end{equation}

where $\mathbf{W}_1$, $\mathbf{W}_2$, $b_1$, and $b_2$ are learnable parameters. The normalized attention weights are then obtained through the softmax operation:

\begin{equation}
\alpha_t =
\frac{\exp(e_t)}
{\sum_{k=1}^{T}\exp(e_k)}
\end{equation}

The attention mechanism assigns higher weights to discriminative temporal segments, improving classification accuracy. The weighted context vector and final prediction are:
\vspace{-2mm}
\begin{equation}
c = \sum_{t=1}^{T} \alpha_t h_t
\end{equation}
\vspace{-5mm}
\begin{equation}
y = \mathrm{softmax}(\mathbf{W}_c c + b_c)
\end{equation}
\vspace{-1mm}

\noindent where $c$ is the aggregated temporal representation and $y$ the predicted class distribution.

To jointly optimize the temporal modeling capability of the \gls{LSTM} network and the discriminative feature aggregation of the attention mechanism, a weighted cross-entropy loss with label smoothing was employed to address class imbalance and improve model generalization. The loss function is defined as:

\begin{equation}
\mathcal{L}
=
-
\sum_{i=1}^{C}
w_i \, y_i \log(\hat{y}_i)
\end{equation}

where $C$ denotes the number of sign classes, $w_i$ is the class weight, $y_i$ is the ground-truth label, and $\hat{y}_i$ represents the predicted probability.

Label smoothing was additionally applied during training to improve generalization and mitigate overfitting.

After translating sign language into gloss sequences, large language models (LLMs) were employed to convert the gloss representations into natural language instructions for the VLA system.

\subsection{Implementation of VLA}

After sign recognition, SignVLA connects the front-end perception module to the downstream GR00T policy through a sign-to-language interface. Predicted sign labels are accumulated over a temporal window; a command is committed only when the same class is the argmax for 3 consecutive frames and an 8-frame cooldown has elapsed, preventing repetition artefacts. The committed gloss sequence is mapped to a natural-language instruction (e.g.\ ``pick up the butter and place it in the basket'') using predefined templates, which is then passed to the GR00T policy as its language condition.

The confirmed instruction is used as the language condition for the downstream GR00T policy. At each control step $t$, the environment provides the current visual observation $\mathbf{o}_t$, and the policy predicts an action conditioned on $\mathbf{o}_t$ and the confirmed sign-derived instruction $l$:
\begin{equation}
    \mathbf{a}_t = \pi_{\mathrm{GR00T}}(\mathbf{o}_t, l),
\end{equation}
where $\pi_{\mathrm{GR00T}}$ denotes the downstream GR00T policy and $\mathbf{a}_t$ is the predicted robot action. During task execution, the same confirmed instruction is maintained unless a new sign command is explicitly confirmed. In this way, SignVLA provides a lightweight sign-language interface for GR00T-based robotic manipulation without requiring the model to directly interpret sign-language video.

\section{Experimental Setup}

\subsection{Dataset}

The \gls{LSTM} classifier is trained on \gls{ASL} Citizen~\cite{desai2023asl}, filtered to a 33-sign vocabulary covering \gls{LIBERO}-Object objects and manipulation actions (train: 921 / val: 224 / test: 288 samples). Each sample is represented as a variable-length sequence of shape $(T,544)$, where $T$ denotes the number of frames. Each frame consists of a 544-dimensional feature vector formed by concatenating absolute and wrist-relative 3D landmark coordinates, finger direction vectors, and frame-to-frame velocity differences for both hands, with zero-padding applied when a hand is absent. Data augmentation applies Gaussian noise and temporal masking to improve robustness.

\subsection{Implementation of Sign Language Translation}

\begin{figure}[H]
    \centering
    \includegraphics[width=0.65\linewidth]{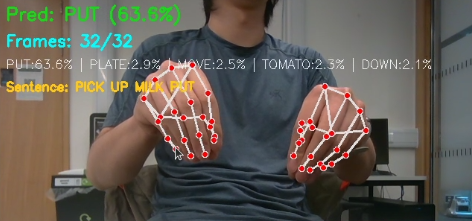}
    \caption{Real-time sign language inference and gesture prediction results.}
    \label{fig:signlan_detec}
\end{figure}

The generated language instructions are designed based on the LIBERO-Object manipulation benchmark. After sign recognition, the predicted sign labels are converted into corresponding gloss tokens and further mapped into natural-language robotic instructions for the downstream VLA policy. A real-time sign recognition example is shown in Figure~\ref{fig:signlan_detec}.

\subsection{GR00T Manipulation Evaluation Setup}

The GR00T policy is supervised fine-tuned on LIBERO demonstration trajectories covering object-to-basket, object-to-container, object-to-surface, and scene interaction tasks (20,000 steps, batch size 640). Each task is evaluated over 20 episodes conditioned on the instruction template associated with a confirmed sign command, yielding 200 total evaluation episodes.

\section{Results}

\subsection{Sign Language Recognition Results}
\subsubsection{Ablation Study on Attention Mechanism and Vocabulary Design}

We evaluate the proposed Attention-LSTM on isolated sign language recognition under two vocabulary settings: a general daily-use vocabulary and a compact LIBERO-Object vocabulary. Table~\ref{tab:ablation_attention} compares the proposed model with a vanilla LSTM baseline. The additive attention mechanism consistently improves recognition performance across both vocabulary settings.

For the general vocabulary setting, the proposed Attention-LSTM improves top-1 accuracy from $25.93\%$ to $67.01\%$. Under the LIBERO-Object vocabulary, the Attention-LSTM achieves $88.89\%$ top-1 accuracy and $96.53\%$ top-5 accuracy, substantially outperforming the vanilla LSTM baseline. These results indicate that additive attention improves temporal motion classification and reduces inter-class ambiguity, especially when the vocabulary is aligned with object-centric robotic manipulation tasks.

\begin{table}[t]
\centering
\caption{Top-1 and Top-5 accuracy (\%) on different vocabularies.}
\label{tab:ablation_attention}
\small
\setlength{\tabcolsep}{5pt}
\begin{tabular}{lcc}
\toprule
\textbf{Model} & \textbf{Top-1} & \textbf{Top-5} \\
\midrule
LSTM (200-word) & 25.93 & 60.36 \\
LSTM (LIBERO)   & 34.72 & 75.35 \\
\midrule
Attn-LSTM (200-word) & 67.01 & 89.23 \\
Attn-LSTM (LIBERO)   & \textbf{88.89} & \textbf{96.53} \\
\bottomrule
\end{tabular}
\end{table}

\subsubsection{Per-Class Recognition Analysis}

Table~\ref{tab:lstm_perclass} reports per-class recognition accuracy for the 33-sign Attention-LSTM on the 288-sample test split. The model achieves strong overall class discrimination, with thirteen signs recognized perfectly, including the majority of task-critical object and action glosses required for downstream \gls{LIBERO} manipulation tasks.

\begin{figure}[t]
\centering
\includegraphics[width=0.75\columnwidth]{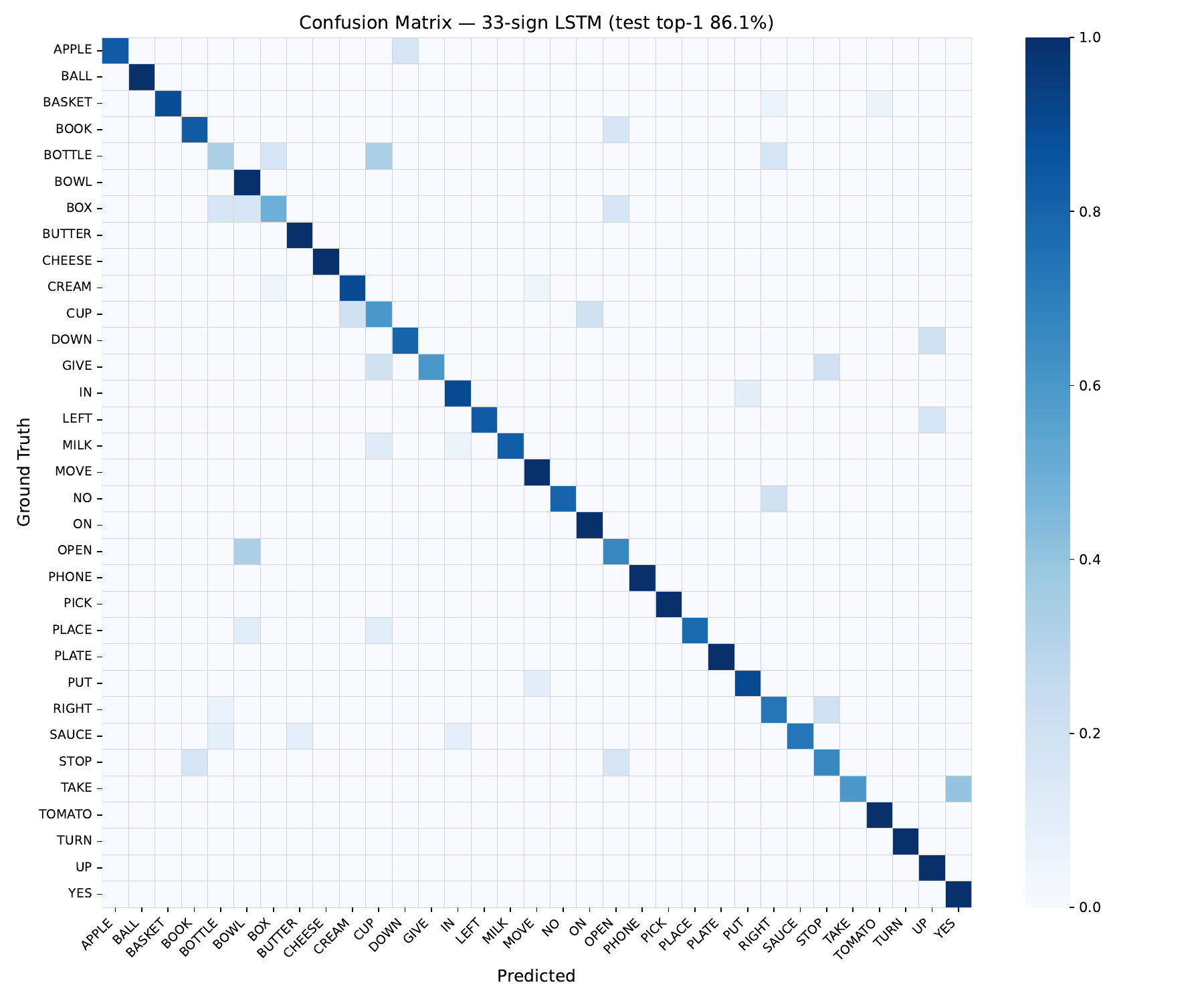}
\caption{Confusion matrix for the 33-sign Attention-\gls{LSTM} on the test set (288 samples). The dominant diagonal confirms strong per-class discrimination; principal off-diagonal mass concentrates at \textsc{bottle}/\textsc{box}/\textsc{bowl} (visually similar handshapes) and \textsc{give}/\textsc{take}/\textsc{stop} (similar motion arcs).}
\label{fig:confusion}
\end{figure}

Residual errors are sparse and concentrated among visually or temporally similar signs rather than uniformly distributed across the vocabulary. As shown in Fig.~\ref{fig:confusion}, most misclassifications occur between object-centric handshape variants or motion-similar action signs, while the confusion matrix remains strongly diagonal overall. This indicates that the remaining errors arise primarily from fine-grained gesture ambiguity rather than broad failure of the recognition model.

\begin{table}[t]
\centering
\caption{Per-class test top-1 accuracy for the 33-sign Attention-LSTM (288 test samples).}
\label{tab:lstm_perclass}
\scriptsize
\setlength{\tabcolsep}{2pt}
\begin{tabular}{lrr|lrr|lrr}
\toprule
\textbf{Sign} & \textbf{C/T} & \textbf{Acc} &
\textbf{Sign} & \textbf{C/T} & \textbf{Acc} &
\textbf{Sign} & \textbf{C/T} & \textbf{Acc} \\
\midrule
APPLE   &  5/6  &  83.3 & DOWN  &  4/5  &  80.0 & PLACE  &  7/9  &  77.8 \\
BALL    &  5/5  & 100.0 & GIVE  &  3/5  &  60.0 & PLATE  & 10/10 & 100.0 \\
BASKET  & 16/18 &  88.9 & IN    &  9/10 &  90.0 & PUT    &  9/10 &  90.0 \\
BOOK    &  5/6  &  83.3 & LEFT  &  5/6  &  83.3 & RIGHT  & 11/15 &  73.3 \\
BOTTLE  &  2/6  &  33.3 & MILK  & 14/17 &  82.4 & SAUCE  &  8/11 &  72.7 \\
BOWL    & 11/11 & 100.0 & MOVE  &  6/6  & 100.0 & STOP   &  4/6  &  66.7 \\
BOX     &  3/6  &  50.0 & NO    &  4/5  &  80.0 & TAKE   &  3/5  &  60.0 \\
BUTTER  & 10/10 & 100.0 & ON    &  9/9  & 100.0 & TOMATO & 12/12 & 100.0 \\
CHEESE  &  9/9  & 100.0 & OPEN  &  4/6  &  66.7 & TURN   &  9/9  & 100.0 \\
CREAM   & 17/19 &  89.5 & PHONE &  5/5  & 100.0 & UP     & 10/10 & 100.0 \\
CUP     &  3/5  &  60.0 & PICK  & 11/11 & 100.0 & YES    &  5/5  & 100.0 \\
\bottomrule
\end{tabular}
\end{table}

\subsection{Sign-Conditioned Robotic Manipulation Results}


Table~\ref{tab:vla_success} reports task success rates obtained using a pretrained GR00T policy on LIBERO task suites under sign-derived language instructions. The sign recognition module converts visual gestures into template-based language commands, which are then used as inputs to the pretrained VLA policy.

\begin{table}[t]
\centering
\caption{GR00T manipulation success rates on LIBERO task suites.}
\label{tab:vla_success}
\small
\setlength{\tabcolsep}{8pt}
\begin{tabular}{lc}
\toprule
\textbf{Task Suite} & \textbf{Success Rate} \\
\midrule
LIBERO-Spatial & 195/200 (97.50\%) \\
LIBERO-Goal & 195/200 (97.50\%) \\
LIBERO-Object & 197/200 (98.50\%) \\
LIBERO-10 (Long) & 189/200 (94.50\%) \\
\bottomrule
\end{tabular}
\end{table}

The pretrained GR00T policy achieves high task success rates across all evaluated LIBERO suites when conditioned on sign-derived instructions generated by the proposed interface.
The highest success rate is observed on LIBERO-Object, suggesting that object-centric manipulation tasks are compatible with the task-constrained sign vocabulary used in this work. Performance on LIBERO-10 is slightly lower, likely due to its longer-horizon execution requirements.

Overall, these results suggest that sign-derived language instructions can serve as a practical low-latency interface for pretrained VLA manipulation policies in simulation.

\begin{figure}[t]
\centering
\includegraphics[width=0.31\linewidth]{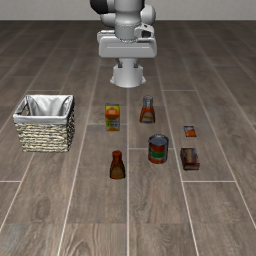}
\includegraphics[width=0.31\linewidth]{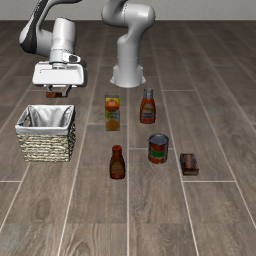}
\includegraphics[width=0.31\linewidth]{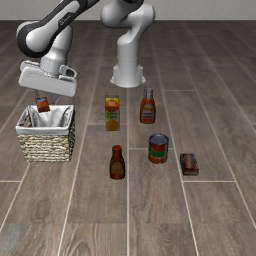}
\vspace{-1mm}
\caption{
Representative rollout of the trained GR00T model on the butter-to-basket instruction.
}
\label{fig:qualitative_pick_butter}
\vspace{-2mm}
\end{figure}

Fig.~\ref{fig:qualitative_pick_butter} shows a representative rollout under a sign-derived butter-to-basket instruction. The trained GR00T model grounds the target object, performs grasping, and completes the placement in the basket.
\subsection{Real-World Franka Deployment Preparation}

\begin{figure}[H]
\centering
\includegraphics[width=0.6\columnwidth]{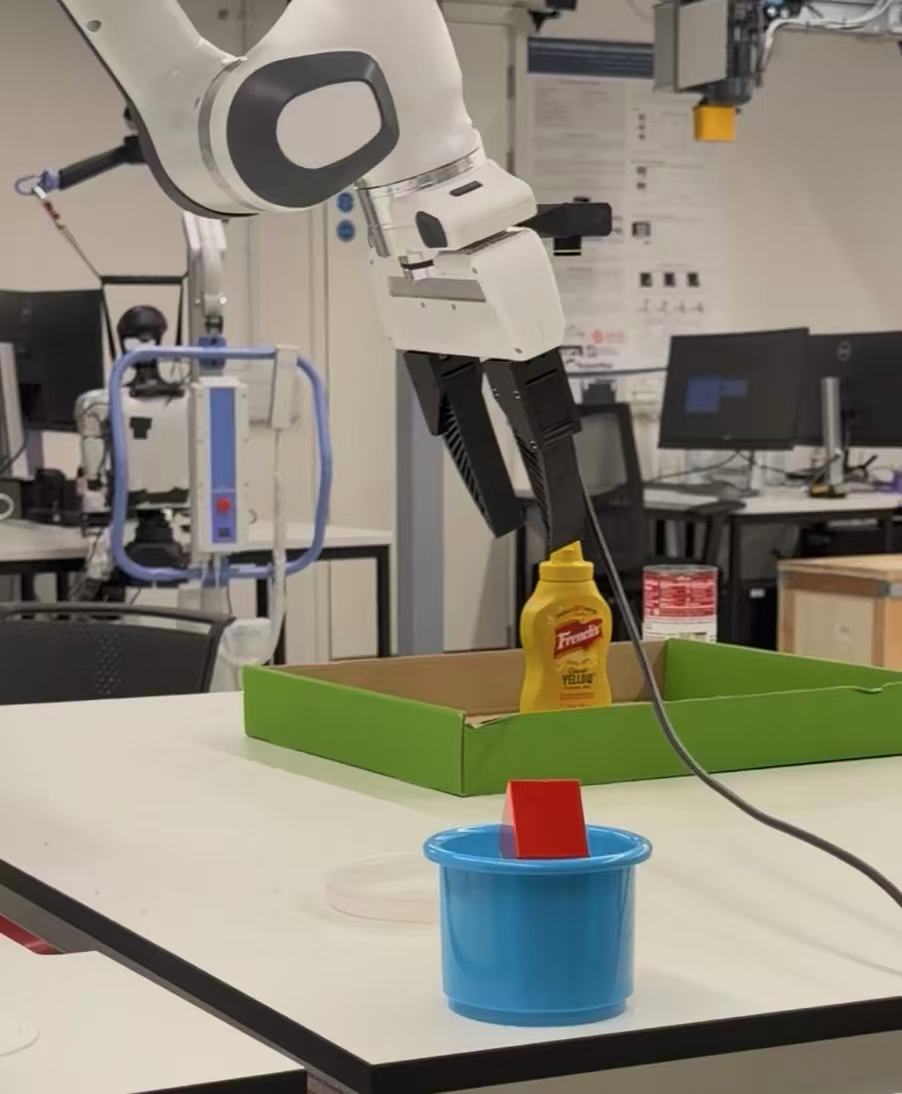}
\caption{
Physical Franka Emika setup prepared for real-world VLA deployment.
The platform will be used for teleoperated demonstration collection, real-robot VLA fine-tuning, and closed-loop manipulation evaluation.
}
\label{fig:real_franka_setup}
\end{figure}

To extend the system beyond simulation, a physical Franka Emika setup has been prepared for real-world VLA deployment, as shown in Fig.~\ref{fig:real_franka_setup}.

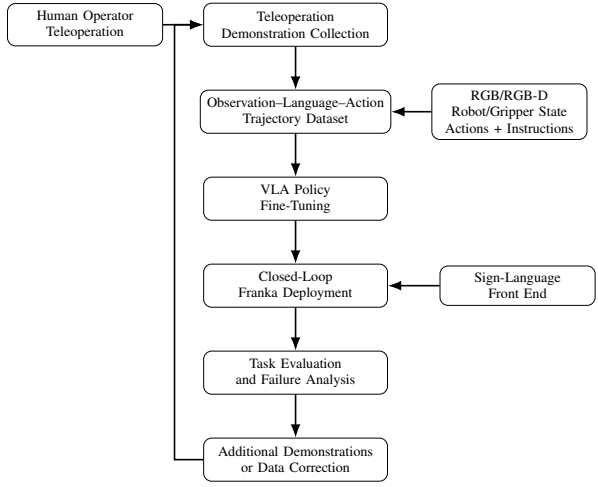
\begin{figure}[H]
\centering
\resizebox{0.9\columnwidth}{!}{
\begin{tikzpicture}[
    node distance=0.75cm,
    block/.style={
        rectangle,
        rounded corners,
        draw,
        align=center,
        minimum width=3.2cm,
        minimum height=0.75cm,
        font=\scriptsize
    },
    sideblock/.style={
        rectangle,
        rounded corners,
        draw,
        align=center,
        minimum width=2.7cm,
        minimum height=0.65cm,
        font=\scriptsize
    },
    arrow/.style={-Latex, thick}
]

\node[block] (teleop) {Teleoperation\\Demonstration Collection};
\node[block, below=of teleop] (dataset) {Observation--Language--Action\\Trajectory Dataset};
\node[block, below=of dataset] (finetune) {VLA Policy\\Fine-Tuning};
\node[block, below=of finetune] (deploy) {Closed-Loop\\Franka Deployment};
\node[block, below=of deploy] (eval) {Task Evaluation\\and Failure Analysis};
\node[block, below=of eval] (augment) {Additional Demonstrations\\or Data Correction};

\node[sideblock, left=0.7cm of teleop] (human) {Human Operator\\Teleoperation};
\node[sideblock, right=0.7cm of dataset] (datafields) {RGB/RGB-D\\Robot/Gripper State\\Actions + Instructions};
\node[sideblock, right=0.9cm of deploy] (sign) {Sign-Language\\Front End};

\draw[arrow] (human) -- (teleop);
\draw[arrow] (teleop) -- (dataset);
\draw[arrow] (datafields) -- (dataset);
\draw[arrow] (dataset) -- (finetune);
\draw[arrow] (finetune) -- (deploy);
\draw[arrow] (deploy) -- (eval);
\draw[arrow] (eval) -- (augment);

\draw[arrow] (sign.west) -- (deploy.east);

\draw[arrow]
    (augment.west) -- ++(-0.5,0) |- (teleop.west);

\end{tikzpicture}
}
\caption{
Planned real-world VLA fine-tuning loop for physical Franka deployment. 
Teleoperated demonstrations are converted into observation--language--action trajectories and used to fine-tune the VLA policy. 
The fine-tuned policy is deployed on the Franka with sign-derived instructions, while failures guide further data collection or correction.
}
\label{fig:real_world_finetuning_loop}
\end{figure}

Fig.~\ref{fig:real_world_finetuning_loop} summarises the planned real-world fine-tuning loop. Teleoperated demonstrations are collected on the robot and formatted into observation--language--action trajectories. The \gls{VLA} policy is then fine-tuned on this physical robot data and deployed for closed-loop manipulation. Failures guide further data collection or task-specific fine-tuning.

The real-world setup has been prepared; complete physical data collection, fine-tuning, and deployment evaluation remain as future work.

\section{Conclusion}


This paper presented SignVLA, a real-time sign-language-guided Vision-Language-Action framework for accessible robotic manipulation. By integrating an attention-based LSTM sign recognizer with a downstream GR00T policy, the proposed system enables robotic manipulation from sign-derived language instructions without modifying the VLA backbone. Experimental results demonstrate strong recognition performance and reliable execution across LIBERO manipulation tasks, suggesting that task-constrained sign vocabularies provide an effective and practical interface for embodied AI systems. Future work will focus on large-vocabulary continuous sign understanding and real-world robotic deployment.

\bibliographystyle{IEEEtran}
\bibliography{ref}

\end{document}